\documentclass[10pt,twocolumn,letterpaper]{article}

\usepackage{cvpr}
\usepackage{times}
\usepackage{epsfig}
\usepackage{graphicx}
\usepackage{amsmath}
\usepackage{amssymb}

\usepackage[breaklinks=true,bookmarks=false]{hyperref}

\cvprfinalcopy 

\def\cvprPaperID{****} 

\begin{document}

\title{A Real-Time Cross-modality Correlation Filtering Method for Referring Expression Comprehension}

\author{Yue Liao\textsuperscript{\rm 1,3}\quad Si Liu\textsuperscript{\rm 1}\thanks{Corresponding author}\quad Guanbin Li\textsuperscript{\rm 2}\quad Fei Wang\textsuperscript{\rm 3}\quad Yanjie Chen\textsuperscript{\rm 3}\quad Chen Qian\textsuperscript{\rm 3}\quad Bo Li\textsuperscript{\rm 1}\\
\large\textsuperscript{\rm 1} School of Computer Science and Engineering, Beihang University \\\quad\textsuperscript{\rm 2} Sun Yat-sen University\quad\textsuperscript{\rm 3} SenseTime Research\\
\tt\small liaoyue.ai@gmail.com; \{liusi, boli\}@buaa.edu.cn; liguanbin@mail.sysu.edu.cn; \\ \tt\small\{wangfei,\ chenyanjie,\ qianchen\}@sensetime.com}

\maketitle
\thispagestyle{empty}
\pagestyle{empty}

\begin{abstract}
Referring expression comprehension aims to localize the object instance described by a natural language expression. Current referring expression methods have achieved good performance. However, none of them is able to achieve real-time inference without accuracy drop. The reason for the relatively slow inference speed is that these methods artificially split the referring expression comprehension into two sequential stages including proposal generation and proposal ranking. It does not exactly conform to the  habit of human cognition.
To this end, we  propose a novel Real-time Cross-modality Correlation Filtering method (RCCF). 
RCCF reformulates the referring expression comprehension as a correlation filtering process.
The expression is first mapped from the language domain to the visual domain and then treated  as a template (kernel) to perform correlation filtering on the image feature map. The peak value in the correlation heatmap indicates the center points of the target box. In addition, RCCF also regresses  a 2-D object size and 2-D offset.
The center point coordinates, object size and center point offset together to form the target bounding box.
 Our method runs at $40$ FPS while achieving leading performance in RefClef, RefCOCO, RefCOCO+ and RefCOCOg benchmarks. In the challenging RefClef dataset, our methods almost double the   state-of-the-art performance  ($34.70\%$ increased to $63.79\%$).  We hope this work can  arouse more attention and studies  to the new cross-modality correlation filtering framework as well as the one-stage framework for referring expression comprehension.
\end{abstract}

\section{Introduction}

\begin{figure}[h]
  \centering
  \includegraphics[width=1\linewidth]{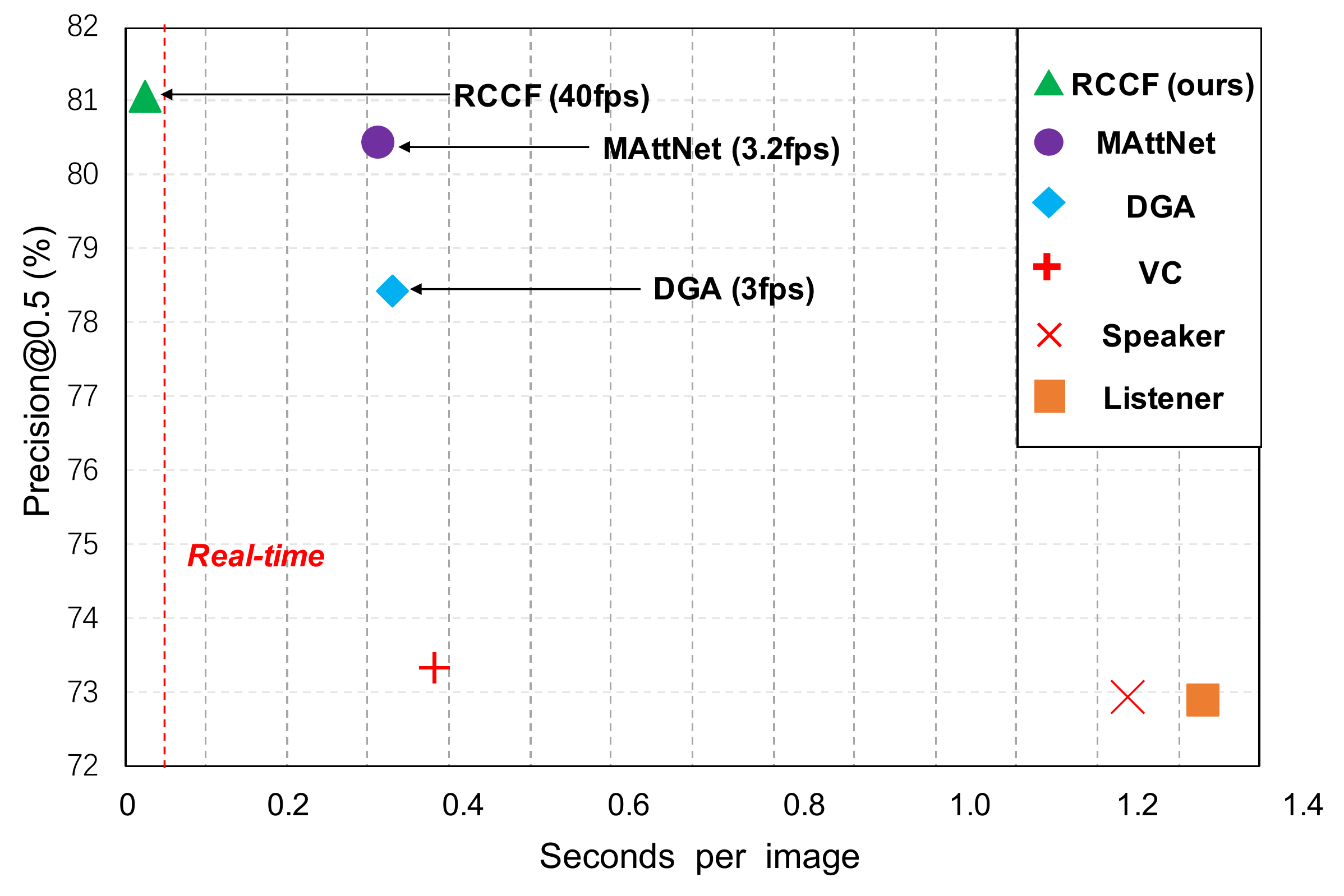}
  \caption{Precision (IOU$>$0.5) versus inference time on the RefCOCO testA set at single Titan Xp GPU. Our method RCCF achieves 40 fps (0.25ms per image), which exceeds the real-time speed of 25 fps and is significantly faster than existing methods by a significant margin (12 times). The precision of RCCF also outperforms the state-of-the-art methods.}
  \label{fig:first}
\end{figure}
  Referring expression comprehension \cite{yu2017joint,yu2018mattnet,wang2019neighbourhood} has attracted much attention in recent years. A referring expression is a natural language description of a particular object in an image. Given such a referring expression, the target of referring expression comprehension is to localize the object instance in the image. It is one of the key tasks in the field of machine intelligence to realize human-computer interaction, robotics and early education.

Conventional methods for referring expression comprehension mostly formulate  this problem as an object retrieval task, where   an object that best matches the referring expression  is retrieved from a set of object proposals. These methods \cite{yu2018mattnet,yang2019dga,yang2019cross,wang2019neighbourhood} are mainly composed of two stages. In the first stage, given an input image, a pre-trained object detection network is applied to generate a set of object proposals. In the second stage, given an input expression,  the best matching region from the detected object proposals is selected.  
Although existing two-stage methods have achieved great advance, there are still some  problems.
1)  The performance of the two-stage methods is very limited to the quality of object proposals generated in the first stage. If the target object is not accurately detected, it is impossible to match  the language in the second stage. 
  2)  In the first stage, a lot of extra object detection data, i.e., COCO \cite{lin2014microsoft} and Visual Genome \cite{krishna2017visual}, are indispensable  to achieve satisfactory result. 
   3)  Two-stage methods are usually computationally  costly. For each object proposal, both 
   feature extraction and cross-modality similarity computation should be conducted. However, only the proposal with highest similarity is selected finally.  As we can see in Figure~\ref{fig:first},  the accuracy of current  two-stage methods is reasonable while the
inference speed still has a large gap to reach real-time.

 The three aforementioned problems  are difficult to solve in existing  two-stage frameworks.
  We reformulate referring expression comprehension as a cross-modality template matching problem, where the language serves  as the template(filter kernel) and the image feature map is the search space to perform correlation filtering on. 
    Mathematically, referring expression comprehension aims to learn a function $f(z,x)$ that compares an  expression  $z$ to a candidate image $x$  and returns a high score in the corresponding regions. 
    The region is represented by 2-dim center point, 2-dim object size (height and width) and 2-dim offset to recover the discretization error \cite{law2018cornernet,zhou2019objects,duan2019centernet}. 
    Our proposed RCCF is  end-to-end trainable.    The  language embedding is used as correlation filter and applied to the feature map to   produce the heatmap for center point.
For more accurate localization, we compute the correlation map on multi-level image feature and fuse the output maps to produce the final heatmap of object center.
       Moreover, the width, height and offset heatmap are regressed with visual feature only. 
       During inference, the text is first embedded into visual space and then slides on the image feature maps. The peak point in the object center heatmap is selected as the center of the target. The corresponding width, height and offset are collected to form the target bounding box, which is the referring expression comprehension result. 
 
       The advantages of our proposed RCCF method can be summarized as three-folds:
\begin{itemize}
 	 \item The inference speed of our method reaches real-time ($40$ FPS)  with a single GPU, which is $12$-times faster than the two-stage methods.
 	 \item Our method can be trained with referring expression dataset only, with no need for any additional object detection data.  Moreover, our one-stage model can avoid error accumulation from the object detector in traditional two-stage methods.
 	 \item RCCF has achieved the state-of-the-art performance in RefClef, RefCOCO, RefCOCO+ and RefCOCOg datasets.  Especially, in the RefClef dataset,  our  method outperforms the state-of-the-art methods by a significant margin from $34.70\%$ to $63.79\%$, almost double the  performance of the state-of-the-art method.
\end{itemize}

\section{Related Work}
\subsection{Referring Expression Comprehension}
Conventional methods for referring expression  comprehension are mostly composed of two-stage. In the first stage, given an input image, a pre-trained object detection network or an unsupervised method is applied to generate a set of object proposals. In the second stage, given an input expression, the best matching region is selected from the detected object proposals. With the development of deep learning, the two-stage methods has achieved great progress. The most two-stage methods focus on improving the second stage.  Most of them \cite{mao2016generation,hu2017modeling,zhang2018grounding,yu2018mattnet,wang2019neighbourhood,yang2019cross} mainly focus on exploring how to mine context information from the language and image or model the relationship between referents, for example, MAttNet~\cite{yu2018mattnet} proposed a modular attention model to capture multi-modality context information. 

Though existing two-stage methods have achieved pretty-well performance, there are some common problems. Firstly, the performance of two-stage methods is limited to the object detectors. Secondly, these methods waste a lot of time in object proposals generation and features extraction for each proposal. Therefore, we propose to localize the target object directly given an expression with our correlation filtering based method.

\subsection{Correlation Filtering}
The correlation filtering is firstly proposed to train a linear template to discriminate between images and their translations. The correlation filtering is widely used in different areas of computer vision. Object classification \cite{krizhevsky2012imagenet,he2016deep,simonyan2014very} can be seen as a correlation filtering task, where the output image feature vector can be seen as a filter kernel, which performs correlation filtering on the weight matrix of the last multi-layer perceptron.  For single object tracking, which aims to localize an object in a video given the object region in the first frame, the correlation filtering can play a role in comparing the first frame with the rest ones. The early works \cite{bolme2010visual,henriques2014high} in tracking firstly transfer the image into Fourier domain, and perform correlation filtering in Fourier domain. Siamese FC \cite{bertinetto2016fully} proposed to directly learn a correlation layer on the spatial domain, where Siamese FC compares two image features extracted from a Siamese network. 

Inspired by human visual perception mechanism, we believe that the process of performing language based visual grounding can be analogized to the process of filter-based visual response activation. Specifically, people generally comprehend the semantic information of a sentence in a global way, and form a feature template about the sentence description in the mind, then quickly perform attention matching on the image based on the template, wherein the salient region with the highest response value is considered as the target matching region.   To this end, we formulate the problem of referring expression comprehension as a cross-modality correlation filtering process and solve with a single-stage joint optimization paradigm.

\begin{figure*}[htb]
  \centering
  \includegraphics[width=1\linewidth]{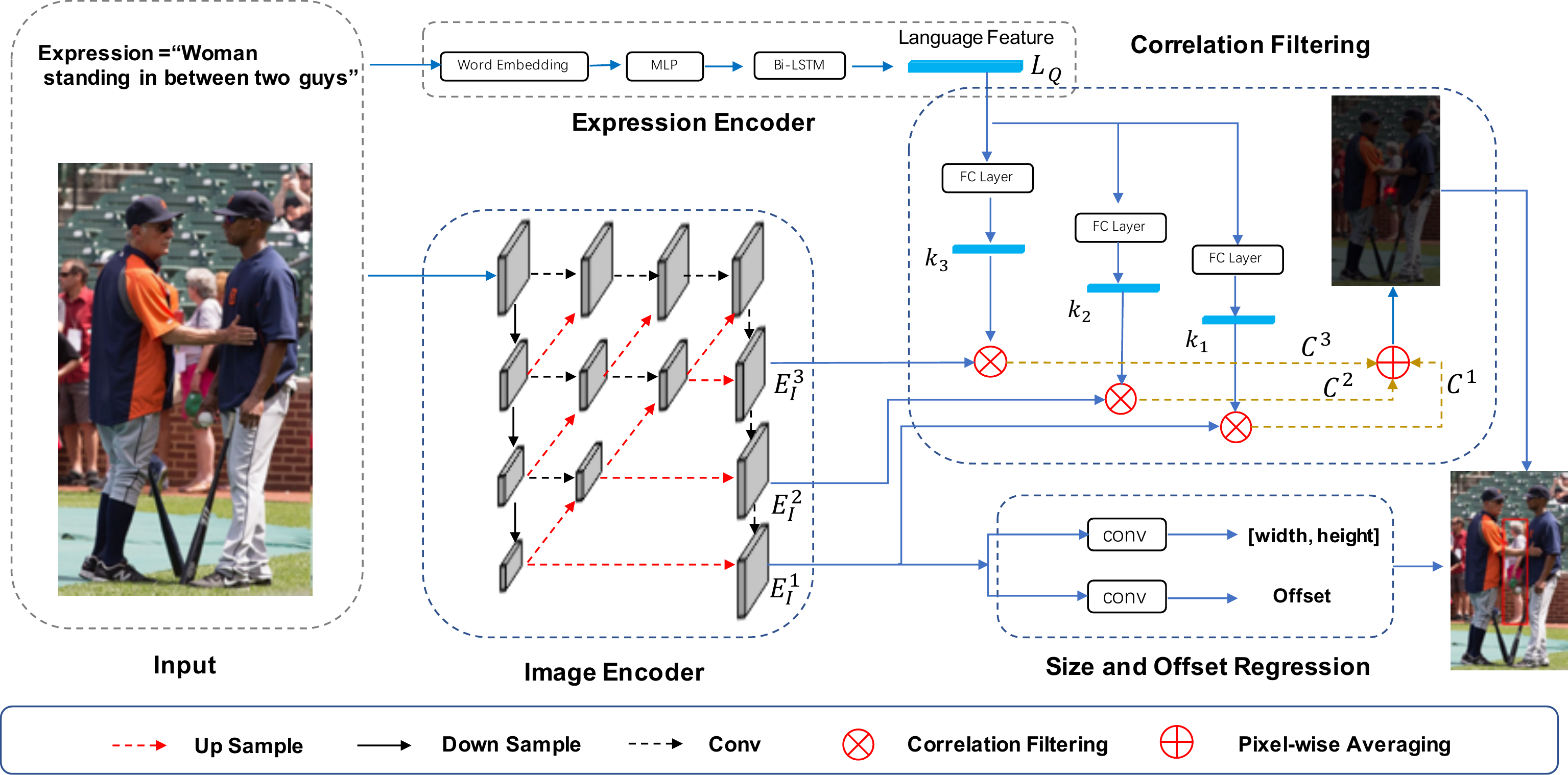}
  \caption{Overview of the proposed RCCF framework. a) Expression and Image Encoder: Bi-LSTM and DLA structure are
    used for expression and visual feature extraction. b) Cross-modality Correlation Filtering: the extracted language feature is mapped into
    three different filter kernels. Then we perform correlation filtering on three levels of image features with the corresponding kernel to generate three correlation maps respectively. Finally, we fuse the three correlation maps by pixel-wise averaging. The center point corresponds to the peak value of the fused heatmap. c) Size and Offset Regression: the 2-dim object size and the local offset for the center point are regressed based on the last-level image feature only. The target object region is obtained by combining the estimated center point, the object size and the local offset.}
  \label{f_work}
\end{figure*}

\section{Method}
In this section, we introduce our proposed RCCF method for referring expression comprehension. Our goal is to localize the object described by the reference expression directly without proposal generation step. To this end, we formulate referring expression comprehension task as a cross-modality template matching problem. In RCCF, we first localize the center point of the object described by the expression by performing correlation filtering on the image feature with a language-guided filter kernel. Then, we apply a regression module to regress the object size and center point offset. The peak value in the correlation heatmap, the regressed object size and center point offset together form the target bounding box.
\subsection{Framework}
\label{sec:pf}
Let $Q$ represent a query sentence and $I \in \mathbb{R}^{H\times W \times 3}$ denote the image of width $W$ and height $H$. Our aim is to find the object region described by the expression. The target object region is represented by  its center point $({x_t}, {y_t})$ and the object size $({w_t}, {h_t})$. Additionally, to recover the discretization error  caused by the output stride, we predict a local offset $ (\delta {x_t}, \delta {y_t})$ for the center point $t$.  To sum up, the referring expression comprehension can
be formulated as a mapping function $({x_t}, {y_t},{w_t}, {h_t},\delta {x_t}, \delta {y_t} )= \phi (Q,I)$.

As shown in Figure~\ref{f_work}, our proposed RCCF is composed of three modules, i.e., expression and image encoder, correlation filtering as well as size and offset regression modules. 
The \emph{expression and image encoder module}  includes the language feature extractor  $L(\cdot)$ and visual feature extractor $E(\cdot)$. The extracted features are represented as $L_Q$ and $E_I$ respectively.  
The expression feature  $L_Q$ is then mapped from the language domain to the visual domain by the cross-modality mapping function $M(\cdot)$.  
The \emph{correlation filtering module}   treats the mapping result $M(L_Q)$  as the  filter (kernel) to convolve with the visual feature map $E_I$ and 
produces a  heatmap $C \in \mathbb{R}^{\frac{H}{d}\times \frac{W}{d}}$, where $d$ is the output stride.
The peak value of $C$  indicates the  center point of the object $({x}, {y})$ depicted by the expression. 
Moreover, the \emph{size and offset  regression module} predicts the object size $({w}, {h})$ and local offset of the center point $(\delta {x}, \delta {y})$.
Next, we will introduce the three modules in detail.

\subsection{Expression and Image  Encoder}

The expression encoder $L(\cdot)$ takes the expression as input, and produces a 512-D feature vector.
We first embed the expression into a 1024-D vector, followed by a fully connected layer to transform the vector into 512-D. Then we feed the transformed feature into a Bi-LSTM to get the expression feature $L_Q$.

The image encoder $E(\cdot)$ adopts the Deep Layer Aggregation (DLA) \cite{yu2018deep} architecture with deformable convolution \cite{dai2017deformable}. DLA is an image
classification network with hierarchical skip connections. Following Centernet \cite{zhou2019objects}, we use the modified DLA network with 34 layers, which replace the skip connection with the deformable convolution.  
 Because a referring expression may consist  of various kinds of semantic information such as  attribute, relationship and spatial location. To well match  the expression, we use three level visual features.
 As shown in  Figure~\ref{f_work}, we extract three level features $[E_I^1, E_I^2, E_I^3] = E(I)$ from the DLA net which are transformed into a unified size $\frac{H}{d} \times \frac{W}{d}$ from $\frac{H}{8d} \times \frac{W}{8d}$, $\frac{H}{4d} \times \frac{W}{4d}$, and $\frac{H}{2d} \times \frac{W}{2d}$ respectively.  The size of $[E_I^1, E_I^2, E_I^3]$ are all $64 \times \frac{H}{d} \times \frac{W}{d}$. When computing the correlation map $\hat{C}$, all three level features are utilized. During regression process, only $E_I^1$ with the  highest resolution is used for computational efficiency.

\subsection{Cross-modality Correlation Filtering}
The aim of cross-modality correlation filtering is to localize the center of the target box $(x, y)$. 
It contains three steps, including language-guided kernel generation, cross-modality correlation operation and correlation maps fusion. Firstly, we utilize three different linear functions to generate three filter kernels $[k_1,k_2,k_3] = [M_1(L_Q), M_2(L_Q), M_3(L_Q)]$ from the expression feature $L_Q$.
The three fully connected layers $M_1(\cdot)$, $M_2(\cdot)$ and $M_3(\cdot)$ serve as the cross-modality mapping function to project from the expression space to the visual space.
Each kernel is a 64-D feature vector which is then reshaped into a $64\times 1 \times 1$ filter for subsequent operations.
Secondly, we perform correlation operation  on the three levels of visual features with their corresponding language-mapped kernels $[C^1,C^2,C^3] = [k_1*E_I^1,k_2*E_I^2,k_3*E_I^3]$, where $*$ denotes convolution operation. Thirdly, the three correlation maps are  pixel-wisely  averaged and fed into an activation function $\hat{C} = Sigmod(\frac{C^1 + C^2 + C^3}{3})$. 
The size of $\hat{C}$, $C^1$, $C^2$ and $C^3$ are all $\mathbb{R}^{\frac{H}{d} \times \frac{W}{d}}$. 
The location with highest score in $\hat{C}$ is the center point of the target object.

We train the center point prediction network following \cite{law2018cornernet,zhou2019objects}. 
For the ground-truth center point $(\tilde{x}^g, \tilde{y}^g)$, we compute a low-resolution equivalent ${({x^g}, {y^g})}=\lfloor\frac{(\tilde{x}^g, \tilde{y}^g)}{d}\rfloor$ by considering the output stride $d$.  We use the Gaussian kernel  $C_{xy} = \exp \left(-\frac{\left(x-{x^g}\right)^{2}+\left(y-{y^g}\right)^{2}}{2 \sigma_{t}^{2}}\right)$ to splat the ground-truth center point in a heatmap $C \in [0,1]^{\frac{W}{d} \times \frac{H}{d}}$, where  $C_{xy}$ is the value of $C$ at the spatial location  $(x,y)$ and $\sigma_t$ is the standard deviation corresponding to the object size. The training objective is a penalty-reduced pixel-wise logistic regression with focal loss \cite{lin2017focal}:
\begin{equation}
L_{c}= -\sum_{x y }\left\{\begin{array}{cc}{\left(1-\hat{C}_{x y}\right)^{\alpha} \log \left(\hat{C}_{x y }\right)} & {\text { if  } C_{x y}=1} \\ {\left(1-C_{x y}\right)^{\beta}\left(\hat{C}_{x y}\right)^{\alpha}} & {\text { otherwise }} \\ {\log \left(1-\hat{C}_{x y }\right)} \end{array}\right.
\end{equation}
where $\alpha$ and $\beta$ are hyper-parameters of the focal loss. We empirically set $\alpha$ to 2, and $\beta$ to 4 in our experiments.

\subsection{Size and Offset Regression} 
As shown in Figure   \ref{f_work}, the module contains  two  parallel branches. The size regression branch predicts the $\hat{W} \in \mathbb{R}^{\frac{H}{d} \times \frac{W}{d}}$ and $\hat{H} \in \mathbb{R}^{\frac{H}{d} \times \frac{W}{d}}$ while the offset regression branch estimates ${\hat{\Delta} x} \in \mathbb{R}^{ \frac{H}{d} \times \frac{W}{d}}$ and ${\hat{\Delta} y} \in \mathbb{R}^{ \frac{H}{d} \times \frac{W}{d}}$ .
The regressed size and offset maps are pixel-wisely corresponded to the estimated center points heatmap $\hat{C}$.

 Both branches take the visual feature $E_I^1$ as input. 
 The regression is conducted without using any expression features. The reason is that the spatial structure information is important for the regression, adding expression features may destroy the rich spatial information in the visual features. 
Both   size and offset regression branches contain  a $3\times 3$ convolutional layer with ReLU followed by a $1\times 1$ convolutional layer.

$L1$ loss function is used during training. The    object size  loss $L_{s i z e}$ and the local offset regression loss  $ L_{off}$ are defined as:
\begin{equation} 
\begin{split}
L_{s i z e}= \left|\hat{W}_{{x^g} {y^g}}-w^g\right| + \left|\hat{H}_{{x^g} {y^g}}-h^g\right|   \\   
 L_{off}= \left|{\hat{\Delta} {x}}_{{x^g} {y^g}} - \delta {x}^g\right| + \left|{\hat{\Delta} {y}}_{{x^g} {y^g}}- \delta {y}^g\right|,
\end{split}
\end{equation}
where $w^g$ and $h^g$ are the ground truth width and height of the target box and $\delta {x}^g = {(\frac{x^g}{d}- {x^g})}$ and $\delta {y}^g = {(\frac{y_g}{d}- {y^g})}$  are the ground truth offset vector.
$\hat{W}_{{x^g} {y^g}}$ is the value of $\hat{W}$ at the spatial location $({x^g},{y^g})$ while 
$\hat{H}_{{x^g} {y^g}}$, ${\hat{\Delta} {x}}_{{x^g} {y^g}}$ and ${\hat{\Delta} {y}}_{{x^g} {y^g}}$ are defined similarly.
 Note that the regression loss acts only at the location of the center point ${({x^g}, {y^g})}$, all other locations are ignored.

\subsection{Loss and Inference}
 
The final loss is the weighted summation of three loss terms:
\begin{equation}
Loss=L_{c}+\lambda_{s i z e} L_{s i z e}+\lambda_{o f f} L_{o f f}
\end{equation}
where we set $\lambda_{s i z e}$ to 0.1 and $\lambda_{o f f}$ to 1. $\lambda_{size}$ is equivalent to a normalized coefficient for the object size.

During inference, we select the point  $({x_t}, {y_t})$ with the highest confidence score in the heatmap $\hat{C}$ as the target center point.  
The target size and offset are obtained from the corresponding position in the $\hat{W}$, $\hat{H}$, $\hat{\Delta x}$ and $\hat{\Delta y}$ as ${\hat{W}_{{x_t}, {y_t}}}$, ${\hat{H}_{{x_t}, {y_t}}}$, ${\hat{\Delta} x_{{x_t}, {y_t}}}$ and ${\hat{\Delta} y_{{x_t}, {y_t}}}$.
The coordinates of the top-left and bottom-right corner of the target box  are obtained by: 
\begin{equation}
\begin{aligned}
(x_t+{\hat{\Delta} x_{{x_t}, {y_t}}}-\frac{{\hat{W}_{{x_t}, {y_t}}}}{2} , y_t+\hat{\Delta} y_{{x_t}, {y_t}}-\frac{{\hat{H}_{{x_t}, {y_t}}}}{2},  \\
 x_t+{\hat{\Delta} x_{{x_t}, {y_t}}}+\frac{{\hat{W}_{{x_t}, {y_t}}}}{2}, 
y_t+{\hat{\Delta} y_{{x_t}, {y_t}}}+\frac{{\hat{H}_{{x_t}, {y_t}}}}{2}). 
\end{aligned}
\end{equation}

\section{Experiments} 
In this section, we first introduce the experimental setting and implementation details, and then evaluate our method on four public benchmarks comparing to the state-of-the-art methods. After that, we analyze  in detail the effectiveness of each component in our framework through a set of ablation experiments.  Finally, we conduct an efficiency analysis followed by the qualitative results analysis.
\subsection{Experimental Setting}
\vspace{-1.5mm}
\paragraph{Dataset.}
The experiments are conducted and evaluated on four common referring expression benchmarks, including RefClef \cite{kazemzadeh2014referitgame}, RefCOCO \cite{kazemzadeh2014referitgame}, RefCOCO+ \cite{kazemzadeh2014referitgame} and RefCOCOg \cite{mao2016generation}. RefClef is also known as Refitgame, and is a subset of the ImageCLEF dataset. The other three datasets are all built on MS COCO images. RefCOCO and RefCOCO+ are collected in an interactive game, where the referring expressions tend to be short phrases. Comparing to RefCOCO, RefCOCO+ forbids using absolute location words and takes more attention on appearance description. To produce longer expressions, RefCOCOg is collected in a non-interactive setting. RefClef has $130,363$ expressions for $99,296$ objects in $19,997$ images. RefCOCO has $142,210$ expressions for $50,000$ objects in $19,994$ images, RefCOCO+ has $141,565$ expressions for $49,856$ objects in $19,992$ images, and RefCOCOg has $104,560$ expressions for $54,822$ objects in $26,711$ images.

Both RefCOCO and RefCOCO+ are divided into four subsets: `train', `val', `testA' and `testB'. The focus of the `testA' and `testB' are different. An image contains multiple people in `testA' and multiple objects in `testB'.
For RefCOCOg, we follow the split in \cite{yu2018mattnet}. For fair comparison, we used the split released by \cite{zhang2018grounding} for RefClef.
\begin{table}[h]
  \begin{center}

  \begin{tabular}{|c|c|c|c|}
  \hline
              & \multicolumn{1}{c|}{\begin{tabular}[c]{@{}c@{}}Params \\ (Million)\end{tabular}} & \begin{tabular}[c]{@{}c@{}}FLOPs \\ (Billion)\end{tabular} & \begin{tabular}[c]{@{}c@{}}Top-1 Error \\ (\%)\end{tabular} \\ \hline
  VGG16      & 138                                   & 15.3            & 28.07            \\  
  ResNet-101 &        44.5                               & 7.6             & 21.75            \\ 
  DLA-34     & \multicolumn{1}{c|}{18.4}             & 3.5             & 25.32            \\ \hline
  \end{tabular}
  \end{center}
  \caption{The parameters, computation and top-1 error on ImageNet validation of the three backbone networks used in referring expression comprehension methods.}
  \label{tb:backbone}
  \end{table}

  \vspace{-1.5mm}
\paragraph{Evaluation Metric.} Following the detection proposal setting in the previous works, we use the Prec@0.5 to evaluate our method, where a predicted region is correct if its intersection over union (IOU) with the ground-truth bounding box is greater than $0.5$.

\subsection{Implementation Details}
We set hyper-parameters following Centernet \cite{zhou2019objects}. Our RCCF method is also robust to these hyper-parameters. All experiments are conducted on the Titan Xp GPU and CUDA 9.0 with Intel Xeon CPU E5-2680v4@2.4G.

The resolution of the input image is $512 \times 512$, and we set the output stride to $4$. Thereby the output resolution is $128 \times 128$.
Our proposed model is trained with Adam \cite{kingma2014adam}. We train on 8 GPUs with a batch-size of $128$ for $80$ epochs, with a learning rate of
5e-4 which is decreased by $10$ at the $60$ epochs, and again at $70$ epochs.  We use random shift and random scaling as the data augmentation. There is none augmentation during inference. The visual encoder are initialized with the weights pretrained on COCO’s training images excluding the val/test set of RefCOCO series datasets, and the language encoder and the output heads are randomly initialized. For ablation study, we also conduct experiments on the visual encoder initialized with ImageNet \cite{deng2009imagenet} pretrain.

\begin{table}[h]
  \begin{center}
  \begin{tabular}{c|c}
  \hline
  Method  & Precise@0.5 (\%) \\ \hline
  SCRC \cite{hu2016natural}    & 17.93    \\ 
  GroundR \cite{rohrbach2016grounding} & 26.93    \\ 
  MCB \cite{chen2017msrc}    & 26.54    \\ 
  CMN \cite{hu2017modeling}    & 28.33    \\ 
  VC \cite{zhang2018grounding}     & 31.13    \\
  GGRE \cite{luo2017comprehension}   & 31.85    \\ 
  MNN  \cite{chen2017msrc}   & 32.21    \\ 
  CITE \cite{plummer2018conditional}   & 34.13    \\
  IGOP \cite{yeh2017interpretable}   & 34.70    \\ 
   \hline
     Ours     &   \textbf{63.79}       \\ \hline
  \end{tabular}
  \end{center}
  \caption{Comparison with the state-of-the-arts  on RefClef.}
  \label{tb:clef}
  \end{table}

\subsection{Comparison to the State-of-the-art}
We compare RCCF to the state-of-the-art methods on four public benchmarks. The comparison results on  RefClef dataset is  shown  in Table~\ref{tb:clef} while the  results on the other three dataset are illustrated in Table~\ref{tb:sota}. The previous methods use a 16-layer VGGNet \cite{simonyan2014very} or a 101-layer ResNet \cite{he2016deep} as the image encoder, while our proposed RCCF adopts DLA-34 \cite{yu2018deep}   to encode images. The reason is that the VGG16 and ResNet-101   are not suitable for the keypoint estimation alike tasks  according to~\cite{law2018cornernet,duan2019centernet} . \begin{table*}[ht]
\small
  \begin{center}
  \begin{tabular}{|c|c|c|c|c|c|c|c|c|c|}
  \hline
                               &          &     & \multicolumn{2}{c|}{RefCOCO} & \multicolumn{2}{l|}{RefCOCO+} & \multicolumn{1}{l|}{RefCOCOg}& \\ \hline
                          &  Method   & Visual Encoder           & testA   & testB     & testA    & testB   & test& Time (ms)                    \\ \hline 
  1 & MMI \cite{mao2016generation}                          & VGG16         &     64.90   & 54.51   &    54.03    & 42.81   &         -        & -   \\
  2 & NegBag \cite{nagaraja2016modeling}                      & VGG16         &   58.60   & 56.40     &    -      &    -           & 49.50     & -    \\
  3 & CG \cite{luo2017comprehension}                          & VGG16         &     67.94   & 55.18      & 57.05    & 43.33   &           -     &-      \\
  4 & Attr \cite{liu2017referring}                         & VGG16         &  72.08   & 57.29      & 57.97    & 46.20   &         -        &-   \\
  5 & CMN \cite{hu2017modeling}                         & VGG16              & 71.03   & 65.77         & 54.32    & 47.76         &            -  &- \\
  6 & Speaker \cite{yu2016modeling}                     & VGG16              & 67.64   & 55.16         & 55.81    & 43.43        &      -       &-  \\
  7 & \textbf{Speaker}+Listener+Reinforcer \cite{yu2017joint} & VGG16          & 72.94   & 62.98       & 58.68    & 47.68        &      -      &1235   \\
  8 & Spearker+\textbf{Listener}+Reinforcer \cite{yu2017joint} & VGG16            & 72.88   & 63.43       & 60.43    & 48.74   &          -       &1332  \\
  9 & VC\cite{zhang2018grounding}                  & VGG16              & 73.33   & 67.44    & 58.40    & 53.18         &   -       &383     \\
  
  10 & ParallelAttn \cite{zhuang2018parallel}               & VGG16             & 75.31   & 65.52   &      61.34    & 50.86   &        -       &-          \\
  11 & LGRANs \cite{wang2019neighbourhood}                       & VGG16         & 76.6    & 66.4      & 64.0     & 53.4    &        -        &-      \\
  12 & DGA \cite{yang2019dga}                       & VGG16         & 78.42    & 65.53      & 69.07     & 51.99    &       63.28          &330  \\ \hline
  13 & \textbf{Spearker}+Listener+Reinforcer \cite{yu2017joint} & ResNet-101        & 73.71   & 64.96       & 60.74    & 48.80           & 59.63    &-     \\
  14 & Spearker+\textbf{Listener}+Reinforcer \cite{yu2017joint} & ResNet-101      & 73.10   & 64.85     & 60.04    & 49.56         & 59.21   &-      \\
  15 & MAttNet \cite{yu2018mattnet}     & ResNet-101   & 80.43   & 69.28    & 70.26    & 56.00    & \textbf{67.01}  & 314       \\ \hline 
          16 &  Ours                 & DLA-34              &  \textbf{81.06}       &    \textbf{71.85}              &   \textbf{70.35}       &   \textbf{56.32}      &       65.73            &25       \\  \hline
  \end{tabular}
  \end{center}
  \caption{Comparison with state-of-the-art approaches on RefCOCO, RefCOCO+ and RefCOCOg.}
  \label{tb:sota}
  \end{table*}
  
For fair comparison, we compare the two backbone networks with  DLA-34 from three aspects in Table~\ref{tb:backbone}. We can see the DLA-34 has the minimum  parameters and computations (FLOPs), and its performance in image classification on ImageNet~\cite{deng2009imagenet} is worse than ResNet-101. 

Therefore, the performance gain of our RCCF comes from the framework itself, instead of more parameters or more complex backbone network. 
The baselines we compared with mainly use Faster-Rcnn \cite{ren2015faster}, pretrained in object detection dataset, i.e., COCO and Visual Genome, to generate object proposals first, then matches the expression with all object proposals.

\vspace{-1.5mm}
\paragraph{RefClef.} 
The results in RefClef are presented in Table~\ref{tb:clef}. Comparing to the state-of-the-art methods in RefClef, our method increases the state-of-the-arts by a significant margin from 34.70\% to 63.79\%, almost double the precision. 

\vspace{-1.5mm}
\paragraph{RefCOCO, RefCOCO+ and RefCOCOg.} Refer to Table~\ref{tb:sota}, our method outperforms existing methods in all evaluation sets on RefCOCO and RefCOCO+, and achieves comparable performance with the state-of-the-art method on RefCOCOg. 
Our result is a slightly inferior to MAttNet \cite{yu2018mattnet} in the RefCOCOg dataset. The performance gain of MAttNet partly comes from the  additional supervision, such as attributes and class labels of region proposals, while our method only utilizes the language-image pair. Additionally, MAttNet uses a more complex  backbone ResNet-101 while we only use DLA-34.

In conclusion, our method can achieve pretty-well performance  in all of the four datasets. In addition, the two-stage methods achieve much higher precision in the three RefCOCO series datasets than in RefClef. It is owing that all three RefCOCO series datasets are  subsets of COCO, so the two-stage methods can train a very accurate detector based COCO object detection dataset, while RefClef does not have a such large corresponding object detection dataset. Therefore,  traditional two-stage methods are heavily  dependent on the object detector performance and the object detection dataset, while our novel RCCF framework  avoid  the explicit object detection stage and tackles the referring expression  problem straightly.

\subsection{Ablation Studies}
In this section, we perform ablation studies from five different aspects on RefCOCO dataset to analyse the rationality and effectiveness of the proposed components in RCCF. The results are shown in Table \ref{tb:ab}.

\vspace{-1.5mm}
\paragraph{Fusion Strategy.} In the first two rows, we report the results on two different fusion manners for the output correlation maps. In the first manner, we fuse the correlation by pixel-wisely taking the maximum value. To accomplish it, we concatenate the three output correlation maps, and obtain pixel-wise maximum across all channels. In the second manner, we generate the output heatmap by concatenating the three correlation maps, followed by a $1 \times 1$ convolutional layer. The results can be seen in the first row and the second row in Table~\ref{tb:ab}. We conclude both the maximum  fusion  and concatenation  are not as good as the average fusion shown in row 10. 
\begin{figure*}[h]
  \centering
  \includegraphics[width=1\linewidth]{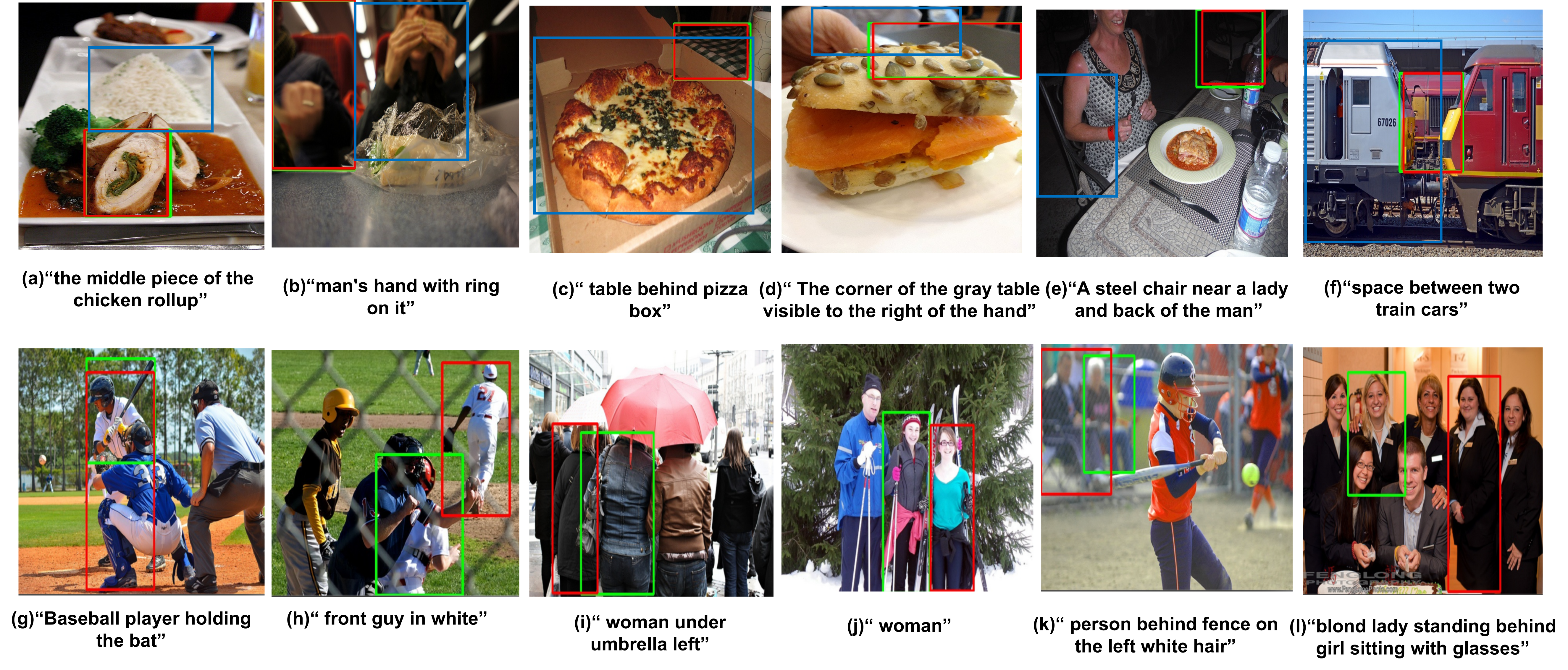}
  \caption{Visualization results  on RefCOCO series dataset.  The first row (a-f) shows the comparisons of our approach with the state-of-the-art method MAttNet. The second row shows some representative failure cases of our method.  The red bounding-box represents the prediction of our method, the blue bounding-box represents the prediction of MAttNet, and the green bounding-box is the corresponding ground-truth. }
  \label{fig:vis_all}
\end{figure*}

\vspace{-1.5mm}
\paragraph{Filter Kernel Setting}Here we perform  ablation studies on the different variations of language filters (kernels).  
$3 \times 3$ Filter (row 3) is the method by expanding the language filter channels by $9$ times, and reshaping it into $3 \times 3$.  Then, we perform correlation filter using the $3 \times 3$ kernels. The result is almost the same with  the `Ours' with the $1\times 1$ kernel (row 10). Considering the additional computational cost, we choose to use $1\times 1$ kernel. 
 
 In row 4, we only generate one filter from the language feature, and perform correlation filtering on the three level visual features with the same kernel. In this case, the precision has dropped about 3 points. This shows that the diversity of the language kernels is important to match the visual features of different levels. 

 \vspace{-1.5mm}
 \paragraph{Single Level Visual Feature.} In  row 5, we perform the correlation filtering only based on the last level of the visual feature $E_I^1$ with single language kernel. The performance has dropped a lot from "Ours", but only dropped a little from the single language  filter, multi-level visual features setting in row 4. Therefore, it can be concluded that the different language filters are sensitive to the different level of visual features.

 \vspace{-1.5mm}
 \paragraph{Language-guided Regression.} To verify whether the feature filtered by the language filter is  suitable for the regression, we  feed the concatenated feature of the three correlation maps into  two convolutional layers in two regression branches. As shown in  row 6, the performance drops a lot, about 6 points. Therefore, it is not a good choice to use language-guided features to regress the object size and offset in our RCCF framework.

 \vspace{-1.5mm}
 \paragraph{Expression \& Image Encoder.}
The row 7 to row 9 of Table~\ref{tb:ab} show our method with various encoders. In  row 7, to explore the effect of the visual encoder pretrain model on the performance, we initialize the DLA-34 with ImageNet pretrain instead of COCO object detection pretrain. The results have dropped about 2 points, but also achieved comparable results to the state-of-the-art method. It proves that our method can also work well without any prior knowledge from object detection. 
In row 8, we use GloVe \cite{pennington2014glove} as the word embedding. There is little change in the performance, so our method is robust to the two different language embeddings. 
In  row 9, we replace the visual encoder with a deeper network Hourglass-104 \cite{law2018cornernet} in a single level setting. Comparing to the row 5, this setting has just improved a little, but this setting is much slower than our basic setting with DLA-34 during inference and training.  More than $100$ hours are needed for training and the inference speed is much lower.

\subsection{Efficiency Analysis}
\noindent\textbf{Inference.} As can be seen in Figure~\ref{fig:first}, our model runs at 25ms per image on a single  Titan Xp GPU and is the only real-time method in referring expression comprehension area. In comparison, our method is 12 times faster than the state-of-the-art two-stage method MAttNet which needs to cost 314ms for an image. For more detail comparison, the inference time per image of the first stage and the second stage of MAttNet are 262ms and 52ms respectively. The cost of either stage is longer than the total inference time of our method.
More comparisons of the timing and precision can be found in Figure~\ref{fig:first}.

\begin{table}[h]
\begin{center}
\footnotesize
\begin{tabular}{|c|c|c|c|c|}
\hline
   & \multicolumn{1}{l|}{}          & \multicolumn{2}{c|}{RefCOCO}                                                                                                      &                     \\ \hline
   & \multicolumn{1}{c|}{Method}     & \multicolumn{1}{c|}{testA} & \multicolumn{1}{c|}{testB}  &Time(ms)\\ \hline
1  & Maximum Fusion                                                &                77.16               &                             69.15  &  25                                               \\ 
2  & Concatenation                                          &                        79.85       &             69.83                  &     26                                           \\ \hline
3  & 3x3 Filter                                                     &                    80.83           &           \textbf{72.01}                    &                           26                 \\  
4  & Single Language Filter                             &                           77.66     &               68.87                &                        24                        \\ \hline
5  & Single Level Visual Feature                                      &                     77.14          &            68.50                   &                            23                   \\ \hline
6  & Language-guided Regression                                  &                             75.13  &           66.16                    &                                       24         \\ \hline
7  & ImageNet      Pretrained                                       & 78.93                          &66.73                            & 25                  \\  
8  & Glove Expression Encoder                                          &                     81.05          &             71.17                  &                            25          \\ 
9 & Hourglass  Image Encoder                                              &   78.12                          &               69.38                 &                                           80    \\ \hline
10  &Ours    & \textbf{81.06}    & 71.85      & 25                  \\
\hline
\end{tabular}
\end{center}
\caption{Ablation experiments on RefCOCO dataset.}
\label{tb:ab}
\end{table}
\noindent\textbf{Training.}  Our method is also fast to train. Training with DLA-34 on RefCOCO takes 35 hours in our synchronized 8-GPU implementation  (1.78s per 128 image-language pairs mini-batch).

\subsection{Qualitative Results Analyses}
\begin{figure}[h]
  \centering
  \includegraphics[width=1\linewidth]{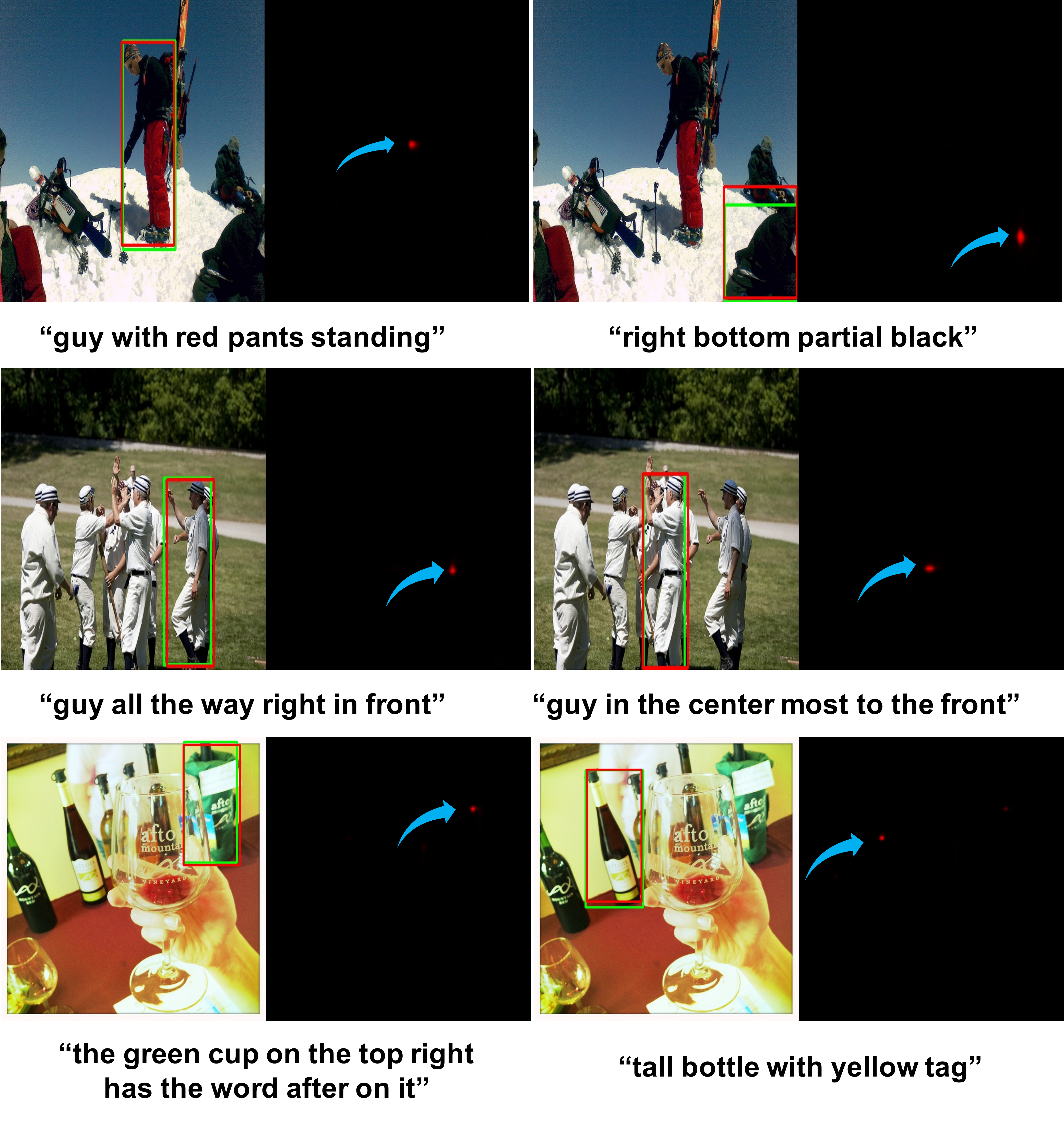}
  \caption{Visualization of visual grounding results and correlation map. On the left image, the red bounding-box represents the prediction of our method while the green bounding-box represents the  ground-truth. The right image shows the  corresponding predicted correlation map for the center point of the object (pointed by the blue arrow).}
  \label{fig:vis_heatmap}
\end{figure}

\vspace{-1.5mm}
\paragraph{Correlation Map.} 
Figure~\ref{fig:vis_heatmap} shows the correlation map of the object center. We can see that given different expressions for the same image, the correlation map  responses to  different locations. 
Otherwise, it can be seen that the  response is very high in areas near the center of object described by the expression. Moreover, there are very small responses in other locations. It shows that our model is capable to well match the expression and   visual features.

\vspace{-1.5mm}
\paragraph{Comparison to the State-of-the-art.} In the first row of  Figure~\ref{fig:vis_all}, we compare our method with the state-of-the-art method MAttNet. Our method can accurately localize the target objects under the guidance of the language, even if the objects are hard to be detected for common object detectors.  For example, although the described objects "piece" (Figure~\ref{fig:vis_all}(a)) and  "space" (Figure~\ref{fig:vis_all}(f))   are very abstract and not included in the COCO categories, our method can still find them through the expression. 
It  proves that our method can well match  expression and visual features.
While MAttNet is dependent on the object detector, MAttNet will fail if the object category is beyond the scope of the detector category set.
\vspace{-1.5mm}
\paragraph{Failure Case Analysis.}
The second row of Figure~\ref{fig:vis_all} illustrates some possible failure cases. As shown in the Figure~\ref{fig:vis_all}(g),  we find the right object, but fail to accurately locate the bounding-box. Another  example is  shown in Figure~\ref{fig:vis_all}(h), the target object is occluded heavily, and the model cannot capture enough appearance information. 
In addition, the ground-truth error may occur.  For example  in Figure~\ref{fig:vis_all}(j), there are more than one target objects described by the expression. 
Some failure cases may be caused by that target object lies in the background and it is difficult to find the appearance feature described by the expression. 
In addition,  when expression is very complex and long,   our model  may fail to understand it well, such as the case in Figure~\ref{fig:vis_all}(l). We leave how to solve these failure cases as interesting future works.

\vspace{-1.5mm}
\section{Conclusion and Future Works}
In this paper, we propose a real-time and high-performance framework for referring expression comprehension. Completely different from the previous two-stage methods, our proposed RCCF directly localizes the object given an expression by predicting the object center through computing a correlation map between the referent and the image. The RCCF is able to achieve state-of-the-art performance in four referring expression datasets at real-time speed. For future work, on the one hand, we plan to explore how to capture more context information from expression and image, and thus understand the expression better. On the other hand, the referring
expression is difficult to annotate, so we want to explore how to utilize other easy annotated types of datasets to train our
method, like object detection, image caption.

\vspace{-1.5mm}
\paragraph{Acknowledgement} This work was partially supported by the State Key Development Program (Grant 2016YFB1001004), Sensetime Ltd. Group, the National Natural Science Foundation of China ( Grant 61876177, Grant 61976250),  Beijing Natural Science Foundation (L182013, 4202034), Zhejiang Lab (No. 2019KD0AB04 ), and  Fundamental Research Funds for the Central Universities.

{\small
\bibliographystyle{ieee_fullname}
\bibliography{egbib}
}

\end{document}